\def\year{2021}\relax
\documentclass[letterpaper]{article} 
\usepackage{aaai21}  
\usepackage{times}  
\usepackage{helvet} 
\usepackage{courier}  
\usepackage[hyphens]{url}  
\usepackage{graphicx} 
\usepackage{natbib}  
\usepackage{caption} 
\usepackage{bm}
\usepackage{dsfont}
\usepackage{mathrsfs}
\usepackage{ amssymb }
\usepackage{ amsopn }
\usepackage{mathtools}
\usepackage{booktabs} 
\usepackage[normalem]{ulem}  
\useunder{\uline}{\ul}{} 
\usepackage{amsthm}  
\newcommand{\Tau}{\mathcal{T}}
\newcommand*{\argdot}{\makebox[1ex]{\textbf{$\cdot$}}}%

\usepackage{fancyvrb} 
\usepackage{ upgreek }

\usepackage{relsize} 

\usepackage{thmtools} %
\usepackage{thm-restate} 

\title{ChronoR: Rotation Based Temporal Knowledge Graph Embedding}
\author {

        Ali Sadeghian\thanks{Authors contributed equally.}\textsuperscript{\rm 1},
        Mohammadreza Armandpour\textsuperscript{\rm *}\textsuperscript{\rm 2},
        Anthony Colas \textsuperscript{\rm 1},
        Daisy Zhe Wang \textsuperscript{\rm 1} \\
}
\affiliations {
    \textsuperscript{\rm 1} University of Florida \\
    \textsuperscript{\rm 2} Texas A\&M University \\
    \{asadeghian, acolas1, daisyw\}@ufl.edu, armand@stat.tamu.edu
}
\begin{document}

\maketitle

\begin{abstract}
Despite the importance and abundance of temporal knowledge graphs, most of the current research has been focused on reasoning on static graphs. In this paper, we study the challenging problem of inference over temporal knowledge graphs. In particular, the task of temporal link prediction. In general, this is a difficult task due to data non-stationarity, data heterogeneity, and its complex temporal dependencies. 
We propose \emph{Chronological Rotation embedding} (ChronoR), a novel model for learning representations for entities, relations, and time. Learning dense representations is frequently used as an efficient and versatile method to perform reasoning on knowledge graphs. The proposed model learns a k-dimensional rotation transformation parametrized by relation and time, such that after each fact's head entity is transformed using the rotation, it falls near its corresponding tail entity. By using high dimensional rotation as its transformation operator, ChronoR captures rich interaction between the temporal and multi-relational characteristics of a Temporal Knowledge Graph. Experimentally, we show that ChronoR is able to outperform many of the state-of-the-art methods on the benchmark datasets for temporal knowledge graph link prediction.
\end{abstract}


\section{Introduction}
\begin{figure}[t!]
    \centering
    \includegraphics[width=0.46\textwidth]{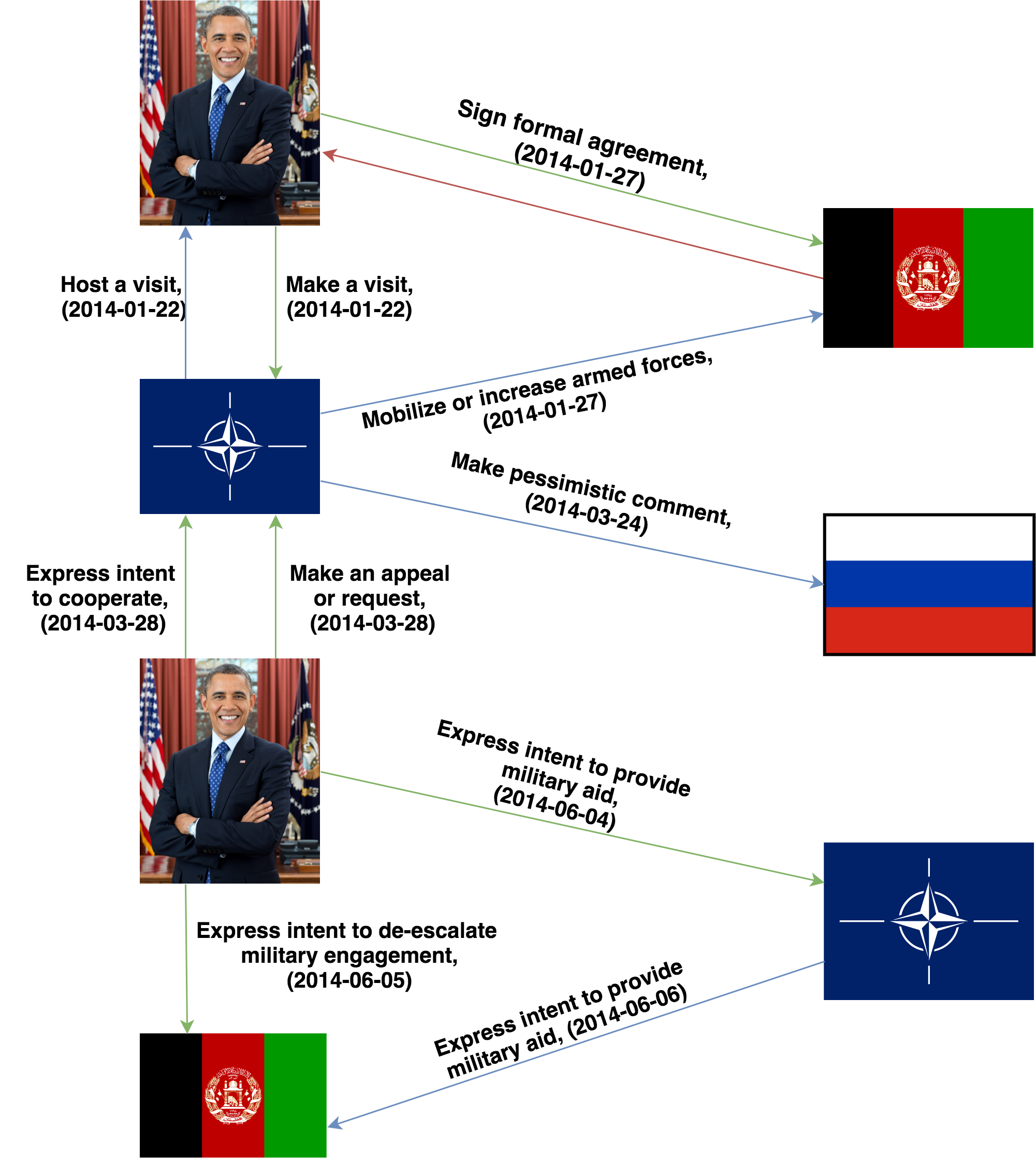}
    \caption{A~timeline of events extracted from the ICEWS14 knowledge graph. The events are chronologically sorted from top to bottom, demonstrating interactions between, president Obama, NATO, Afghanistan~and~Russia.}
    \label{fig:ICEWS14}
\end{figure}

Knowledge Graphs (KGs) organize information around entities (people, countries, organizations, movies, etc.) in the form of factual triplets, where each triplet represents how two entities are related to each other, for example (Washington DC, \emph{captialOf}, USA).

There exists an ever growing number of publicly available KGs, for example DBPedia~\cite{auer2007dbpedia}, Freebase~\cite{bollacker2008freebase}, Google Knowledge Graph~\cite{blog2012introducing}, NELL~\cite{carlson2010toward}, OpenIE~\cite{yates2007textrunner, etzioni2011open}, YAGO~\cite{biega2013inside, hoffart2013yago2, mahdisoltani2013yago3}, and UMLS~\cite{burgun2001comparing}. This structured way of representing knowledge makes it easy for computers to digest and utilize in various applications. For example KGs are used in recommender systems~\cite{cao2019unifying, zhang2016collaborative}, the medical domain~\cite{sang2018sematyp, abdelaziz2017large}, question-answering~\cite{hao2017end}, information retrieval~\cite{xiong2017explicit}, and natural language processing~\cite{yang2017leveraging}.

Though these graphs are continuously growing, they remain particularly incomplete. Knowledge Base Completion (KBC) focuses on finding/predicting missing relations between entities. A wide range of work explores various methods of finding extraction errors or completing KGs~\cite{bordes2013translating, yang2014embedding, trouillon2016complex, sadeghian2019drum, zhou2019mining, sun2019rotate, zhang2019quaternion}.

Different events and actions cause relations and entities to evolve over time. For example, Figure~\ref{fig:ICEWS14} illustrates a timeline of relations and events happening in 2014 involving entities like Obama (president of USA at the time), NATO, Afghanistan, and Russia. Here, an agreement between Obama and Afghanistan is happening concurrently with NATO increasing armed forces in Afghanistan; after Obama made a visit to NATO. A few months later, after NATO makes some pessimistic comments towards Russia and Obama wants to de-escalate the conflict, it appears that he provides military aid to NATO, and in turn, NATO provides military aid in Afghanistan. 

Temporally-aware KGs are graphs that add a fourth dimension, namely time \textit{t}, giving the fact a temporal context. Temporal KGs are designed to capture this temporal information and the dynamic nature of real-world facts. While many of the previous works study knowledge graph completion on static KGs, little attention has been given to temporally-aware KGs. Though recent work has begun to solve the temporal link prediction task, these models often utilize a large number of parameters, making them difficult to train~\cite{garcia2018learning,dasgupta2018hyte,leblay2018deriving}. Furthermore, many use inadequate datasets such as YAGO2~\cite{hoffart2013yago2}, which are sparse in the time domain, or a time augmented version of FreeBase~\cite{bollacker2008freebase}, where time is appended to some existing~facts.

One of the most popular models used to solve the link prediction task involves embedding the KG, that is, mapping entities and relations in the KG to high dimensional vectors, in which each entity and relation mapping considers the structure of the graph as constraints~\cite{bordes2013translating,yang2014embedding,lin2015learning}. These techniques have proven to be the state-of-the-art in modeling static knowledge graphs and inferring new facts from the KG based on the existing ones. Similarly, temporal knowledge graph embedding methods learn an additional mapping for time. Different methods differ based on how they map the elements in the knowledge graph and their scoring function.

In this paper, we propose ChronoR, a novel temporal link prediction model based on $k$-dimensional rotation. We formulate the link prediction problem as learning representations for entities in the knowledge graph and a rotation operator based on each fact's relation and temporal elements. Further, we show that the proposed scoring function is a generalization of previously used scoring functions for the static KG link prediction task. We also provide insights into the regularization used in other similar models and propose a new regularization method inspired by tensor nuclear norms. Empirical experiments also confirm its advantages. Our experiments on available benchmarks show that ChronoR outperforms previous state-of-the-art methods.

The rest of the paper is organized as follows: Section~\ref{sec:Related Work} covers the relevant previous work, Section~\ref{sec:Problem definition} formally defines the temporal knowledge graph completion problem, in Section~\ref{sec:Model} we go over the proposed model's details and learning procedure, Section~\ref{sec:Optimization} details the loss and regulation functions, Section~\ref{sec:Experiments} discusses the experimental setup, and Section~\ref{sec:Conclusion} concludes our work.

\section{Related Work}

\label{sec:Related Work}
\subsection{Static KG Embeddings}
There has been a substantial amount of research in KG embedding in the non-temporal domain. One of the earliest models, TransE~\cite{bordes2013translating}, is a translational distance-based model, which embeds the entities \textbf{h} and \textbf{t}, along with relation \textbf{r}, and maps them through the function: \textbf{h} + \textbf{r} = \textbf{t}. 
There have been several extensions to TransE, including TransH~\cite{wang2014knowledge} which models relations as hyperplanes; TransR~\cite{lin2015learning} which embed entities and relations in separate spaces and TransD~\cite{ji2015knowledge} in which two vectors represent each element in a triple in order to represent the elements and construct mapping matrices.
Other works, such as DistMult~\cite{yang2014embedding}, represent relations as bilinear functions.  ComplEx~\cite{trouillon2016complex} extends DistMult to the complex space. RotatE~\cite{sun2019rotate} also embeds the entities in the complex space, and treats relations as planar rotations. QuatE~\cite{zhang2019quaternion} embeds each element using quaternions. 

\subsection{Temporal Embeddings}

There has been a wide range of approaches to the problem of temporal link prediction~\cite{kazemi2020representation}. A straightforward technique is to ignore the timestamps and make a static KB by aggregating links across different times~\cite{liben2007link}, then learn a static embedding for each entity. There have been several attempts to improve along this direction by giving more weights to the links that are more recent~\cite{sharan2008temporal, ibrahim2015link, ahmed2016efficient, ahmed2016sampling}. In contrast to these methods, \cite{yao2016link} first learn embeddings for each snapshot of the KB, then aggregate the embeddings by using a weighted average of them. Several techniques have been proposed for the choice of weights of the embedding aggregation, for instance, based on ARIMA \cite{gunecs2016link} or reinforcement learning \cite{moradabadi2017novel}.

Some of the other works to extend sequence models to TKG. \cite{sarkar2007latent} employs a Kalman filter to learn dynamic node embeddings. \cite{garcia2018learning} use recurrent neural nets (RNN) to accommodate for temporal data and extend DistMult and TransE to TKG. For each relation, a temporal embedding has been learned by feeding time characters and the static relation embedding to an LSTM. This method only learns dynamic embedding for relations, not entities. Furthermore, \cite{han2020graph} utilize a temporal point process parameterized by a deep neural architecture.

In one of the earliest works to employ representation learning techniques for reasoning over TKGs~\cite{sadeghian2016temporal}, proposed both an embedding method as well as rule mining methods for reasoning over TKGs. Another related work, t-TransE ~\cite{jiang2016encoding}, learns time-based embeddings indirectly by learning the order of time-sensitive relations, such as \textit{wasBornIn} followed by \textit{diedIn}. \cite{esteban2016predicting} also impose temporal order constraints on their data by adding an element to their quadruple \textit{(s,p,o,t:Bool)}, where \textit{Bool} indicates if the fact vanishes or continues after time \textit{t}. However, the model is only demonstrated on medical and sensory data.

Inspired by the success of diachronic word embeddings, some methods have tried to extend them to the TKG problem~\cite{garcia2018learning, dasgupta2018hyte}. Diachronic methods map every (node, timestamp) or (relation, timestamp) pair to a hidden representation. \cite{Goel2020DiachronicEF} learn dynamic embeddings by masking a fraction of the embedding weights with an activation function of frequencies and \cite{xu2019temporal}~embed the vectors as a direct function of time. Two con-current temporal reasoning methods TeRo~\cite{xu2020tero} and TeMP~\cite{wu2020temp} are also included in the empirical comparison table in Section~\ref{sec:Experiments}.

Other methods, like~\cite{ma2019embedding, sadeghian2019hotel2vec, jain2020temporal, lacroix2020tensor}, do not evolve the embedding of entities over time. Instead, by using a representation for time, learn the temporal behavior. For instance, \cite{ma2019embedding}~change the scoring function based on the time embedding and~\cite{lacroix2020tensor} perform tensor decomposition based on the time representation. 

\section{Problem Definition}
\label{sec:Problem definition}
In this section, we formally define the problem of temporal knowledge graph completion and specify the notations used throughout the rest of the paper.

We represent scalars with lower case letters $a$, vectors and matrices with bold lower case letters $\bm{a}$, higher order tensors with bold upper case letters $\bm{A}$, and the $i$\textsuperscript{th} element of a vector $\bm{a}$ as $a_i$. We use $\bm{a} \circ \bm{b}$ to denote the element wise product of two vectors and $[\bm{A} | \bm{B}]$ to denote matrix or vector concatenation. We denote the complex norm as $|\cdot|$ and $||\cdot||_p$ denotes the vector p-norm; we drop $p$ when $p=2$. 

\vspace{6pt}

A \textbf{Temporal Knowledge Graph} is referred to a set of quadruples $\mathcal{K} = \{(h, r, t, \tau) \mid h,t \in \mathcal{E}, r \in \mathcal{R}, \tau \in \Tau \}$. Each quadruple represents a temporal fact that is true in a world. $\mathcal{E}$ is the set of all entities and $\mathcal{R}$ is the set of all relations in the ontology. The fourth element in each quadruple represents time, which is often discretized. $\Tau$ represents the set of all possible time stamps. 

\vspace{6pt}

\textbf{Temporal Knowledge Graph Completion} (temporal link prediction) refers to the problem of completing a TKGE by inferring facts from a given subset of its facts. In this work, we focus on predicting temporal facts within the observed set $\Tau$, as opposed to the more general problem which also involves forecasting future facts.
\vspace{6pt}

\section{Temporal KG Representation Learning}\label{sec:Model}

In this section, we present a framework for temporal knowledge graph representation learning.
Given a TKG, we want to learn representations for entities, relations, and timestamps (e.g., $\bm{h}, \bm{r}, \bm{t}, \bm{\tau} \in \mathds{R}^{n \times k}$) and a scoring function $g(\bm{h}, \bm{r}, \bm{t}, \bm{\tau}) \in \mathds{R}$, such that true quadruples receive high scores. Thus, given $g$, the embeddings can be learned by optimizing an appropriate cost function.

Many of the previous works use a variety of different objectives and linear/non-linear operators to adapt static KG completion's scoring functions to the scoring function $g(\argdot)$ in the temporal case (see Section~\ref{sec:Related Work}).

One example, RotatE~\cite{sun2019rotate} learns embeddings $\bm{h}, \bm{r}, \bm{t} \in \mathds{C}^n$ by requiring each triplet's head to fall close to its tail once transformed by an (element-wise) rotation parametrized by the relation vector: $\bm{h} \circ \bm{r} = \bm{t}$, where $|r_i| = 1$. Thus, RotatE defines $g(h, r, t) = \lambda - ||\bm{h} \circ \bm{r} - \bm{t}||$, for some $\lambda \in \mathds{R}$.

Our model is inspired by the success of rotation-based models in static KG completion \cite{sun2019rotate, zhang2019quaternion}. For example, to carry out a rotation by an angle $\theta$ in the two dimensional space, one can use the well known Euler's formula $e^{i \theta} = \cos(\theta) + i \sin(\theta)$, and the fact that if $ (x,y) \in \mathds{R}^2$ is represented by its dual complex form $z = x + i y \in \mathds{C}$, then $R_\theta(x,y) \Leftrightarrow  e^{i \theta} z$. We  represent rotation by angle $\theta$ in the 2-dimensional space with $R_\theta(.)$.

\subsection{ChronoR}
\label{sec:ChronoR}

In this paper, we consider a subset of the group of general linear transformations $\operatorname{GL}(k, \mathds{R})$ over the k-dimensional real space, consisting of rotation and scaling and parametrize the transformation by both time and relation. Intuitively, we expect that for true facts:

\begin{equation}
    \operatorname{Q}_{r, \tau}(\bm{h}) = \bm{t}
\end{equation}

\noindent where $\bm{h}, \bm{t} \in \mathds{R}^{n \times k}$ and $\operatorname{Q}_{r, \tau}$ represents the (row-wise) linear operator in k-dimensional space, parametrized by $\bm{r}$ and $\bm{\tau}$. Note that any orthogonal matrix $\bm{Q} \in \mathds{R}^{n \times n}$  (i.e., $\bm{Q}^\mathsf{T} \bm{Q} = \bm{Q} \bm{Q}^\mathsf{T} = I$) is equivalent to a k-dimensional rotation. However, we relax the unit norm constraint, thus extending rotations to a subset of linear operators which also includes scaling.

As previous work has noted, for 2 or 3 dimensional rotations, one can represent $Q$ using complex numbers $\mathds{C}$ and quaternions $\mathds{H}$ ($k = 2 \text{ and } 4$), respectively. Higher dimensional transformations can also be constructed using the Aguilera-Perez Algorithm~\cite{aguilera2004general}
followed by a scalar multiplication.

\subsection{Scoring Function}

Unlike RotatE, that uses a scoring function based on the Euclidean distance of $\operatorname{Q_r}( \bm{h} )$ and $\bm{t}$, we propose to use the angle between the two vectors. 

Our motivations comes from observations in a variety of previous works~\cite{aggarwal2001surprising, zimek2012survey} showing that in higher dimensions, the Euclidean norm suffers from the curse of high dimensionality and is not a good measure of the concept of proximity or similarity between vectors. We use the following well-known definition of inner product to define the angle between $\operatorname{Q}_{r,\tau}(\bm{h})$ and  $\bm{t}$. \\

\begin{restatable}[]{defi}{matrixInnerProduct}
\label{def:matrixInnerProduct}
If $\bm{A}$ and $\bm{B}$ are matrices in $\mathds{R}^{n \times k}$ we define: 

\begin{align}
    \left\langle \bm{A}, \bm{B} \right\rangle &:= \operatorname{tr}(\mathbf{AB^{T}}) \\
    \cos(\theta) &:= \frac{\left\langle \bm{A}, \bm{B} \right\rangle}{ \sqrt{ \left\langle \bm{A}, \bm{A} \right\rangle \left\langle \bm{B}, \bm{B} \right\rangle} } \\ \nonumber
\end{align}
\end{restatable}

Based on the above definition, the angle between two matrices is proportional to their inner product. Hence, we define our scoring function as:
\begin{equation}
\label{eq:scoring_function}
    g(h, r, t, \tau) := \left\langle \operatorname{Q}_{r,\tau}(\bm{h}) \, , \bm{t} \right\rangle
\end{equation}

\noindent That is, we expect the $\cos$ of the relative angle between $\operatorname{Q}_{r,\tau}(\bm{h})$ and $\bm{t}$ to be higher (i.e., their angle close to $0$) when $(h, r, t, \tau)$ is a true fact in the TKG.

Some of the state of the art work in static KG completion, such as quaternion-based rotations QuatE~\cite{zhang2019quaternion}, complex domain tensor factorization~\cite{trouillon2016complex}, and also recent works on temporal KG completion like TNTComplEx~\cite{lacroix2020tensor}, use a scoring function similar to

\begin{align}
    g(h, r, t) &= \operatorname{Re}\{\bm{h} \circ \bm{r} \circ \bm{\bar{t}}\} \\
    \text{for } \, \bm{h} &, \bm{r}, \bm{t} \in \mathds{C} \text{ or } \mathds{H} \nonumber
\end{align}

\noindent and motivate it by the fact that the optimisation function requires the scores to be purely real~\cite{trouillon2016complex}. 

However, it is interesting to note that this scoring method is in fact a special case of Equation~\ref{eq:scoring_function}. The following theorem proves the equivalence for scoring functions used in \emph{ComplEx}\footnote{A similar theorem holds for quaternions when k=4, see Appendix} to Equation~\ref{eq:scoring_function} when $k=2$.

\begin{restatable}[]{thm}{innerProductEquivalence}
\label{thm:innerProductEquivalence}
If $\bm{a}, \bm{b} \in \mathds{C}^n$ and $\bm{A}, \bm{B} \in \mathds{R}^{n \times 2} $ are their equivalent matrix forms, then $\operatorname{Re}(\bm{a} \circ \bm{\bar{b}}) = \left\langle \bm{A}, \bm{B} \right\rangle$ (proof in Appendix)
\end{restatable}

To fully define $g$ we also need to specify how the linear operator $\operatorname{Q}$ is parameterized by $h, \tau$. In the rest of the paper and experiments, we simply concatenate the head and relation embeddings to get $\operatorname{Q}_{r,\tau} = [\bm{r} | \bm{\tau}]$, where $\bm{r} \in \mathds{R}^{n_r \times k}$ and $\bm{\tau} \in \mathds{R}^{n_\tau \times k}$ are the representations of the fact's relation and time elements and $n_r + n_\tau = n$. 
Since in many real world TKGs, there are a combination of static and dynamic facts, we also allow an extra rotation operator parametrized only by $r$, i.e., an extra term $\bm{R}_2 \in \mathds{R}^{n \times k}$ in $\operatorname{Q}$ to better represent static facts.

To summarize, our scoring function is defined as:
\begin{equation}
\label{eq:scoring_function_final}
    g(h, r, t, \tau) := \left\langle \bm{h} \, \circ [\bm{r} | \bm{\tau}] \, \circ \bm{r}_2, \bm{t} \right\rangle
\end{equation} 
where $\bm{r} \in \mathds{R}^{n_r \times k}$, $\bm{\tau} \in \mathds{R}^{n_\tau \times k}$ and $\bm{r}_2 \in \mathds{R}^{n \times k}$. 

\section{Optimization}
\label{sec:Optimization}
\begin{table*}[t]
\resizebox{0.99\textwidth}{!}{%
\begin{tabular}{lcccccccccccc}
\hline
 &
  \multicolumn{4}{c}{{\ul ICEWS14}} &
  \multicolumn{4}{c}{{\ul ICEWS05-15}} &
  \multicolumn{4}{c}{{\ul YAGO15K}} \\
\multicolumn{1}{l|}{Model} &
  MRR &
  Hit@1 &
  Hits@3 &
  \multicolumn{1}{c|}{Hit@10} &
  MRR &
  Hit@1 &
  Hit@3 &
  \multicolumn{1}{c|}{Hit@10} &
  MRR &
  Hit@1 &
  Hit@3 &
  Hit@10 \\ \hline
\multicolumn{1}{l|}{TransE (2013)} &
  28.0 &
  9.4 &
  - &
  \multicolumn{1}{c|}{63.70} &
  29.4 &
  8.4 &
  - &
  \multicolumn{1}{c|}{66.30} &
  29.6 &
  22.8 &
  - &
  46.8 \\
\multicolumn{1}{l|}{DistMult (2014)} &
  43.9 &
  32.3 &
  - &
  \multicolumn{1}{c|}{67.2} &
  45.6 &
  33.7 &
  - &
  \multicolumn{1}{c|}{69.1} &
  27.5 &
  21.5 &
  - &
  43.8 \\
\multicolumn{1}{l|}{SimpIE (2018)} &
  45.8 &
  34.1 &
  51.6 &
  \multicolumn{1}{c|}{68.7} &
  47.8 &
  35.9 &
  53.9 &
  \multicolumn{1}{c|}{70.8} &
  - &
  - &
  - &
  - \\
\multicolumn{1}{l|}{ComplEx (2016)} &
  47.0 &
  35.0 &
  54.0 &
  \multicolumn{1}{c|}{71.0} &
  49.0 &
  37.0 &
  55.0 &
  \multicolumn{1}{c|}{73.0} &
  36.0 &
  29.0 &
  36.0 &
  54.0 \\ \hline
\multicolumn{1}{l|}{ConT (2018)} &
  18.5 &
  11.7 &
  20.5 &
  \multicolumn{1}{c|}{31.50} &
  16.4 &
  10.5 &
  18.9 &
  \multicolumn{1}{c|}{27.20} &
  - &
  - &
  - &
  - \\
\multicolumn{1}{l|}{TTransE (2016)} &
  25.5 &
  7.4 &
  - &
  \multicolumn{1}{c|}{60.1} &
  27.1 &
  8.4 &
  - &
  \multicolumn{1}{c|}{61.6} &
  32.1 &
  23.0 &
  - &
  51.0 \\
\multicolumn{1}{l|}{TA-TransE (2018)} &
  27.5 &
  9.5 &
  - &
  \multicolumn{1}{c|}{62.5} &
  29.9 &
  9.6 &
  - &
  \multicolumn{1}{c|}{66.8} &
  32.1 &
  23.1 &
  - &
  51.2 \\
\multicolumn{1}{l|}{HyTE (2018)} &
  29.7 &
  10.8 &
  41.6 &
  \multicolumn{1}{c|}{65.5} &
  31.6 &
  11.6 &
  44.5 &
  \multicolumn{1}{c|}{68.1} &
  - &
  - &
  - &
  - \\
\multicolumn{1}{l|}{TA-DistMult (2018)} &
  47.7 &
  - &
  36.3 &
  \multicolumn{1}{c|}{68.6} &
  47.4 &
  34.6 &
  - &
  \multicolumn{1}{c|}{72.8} &
  29.1 &
  21.6 &
  - &
  47.6 \\
\multicolumn{1}{l|}{DE-SimpIE (2020)} &
  52.6 &
  41.8 &
  59.2 &
  \multicolumn{1}{c|}{72.5} &
  51.3 &
  39.2 &
  57.8 &
  \multicolumn{1}{c|}{74.8} &
  - &
  - &
  - &
  - \\
\multicolumn{1}{l|}{TIMEPLEX (2020)} &
  60.40 &
  51.50 &
  - &
  \multicolumn{1}{c|}{77.11} &
  63.99 &
  54.51 &
  - &
  \multicolumn{1}{c|}{81.81} &
  - &
  - &
  - &
  - \\
\multicolumn{1}{l|}{TNTComplEx (2020)} &
  60.72 &
  51.91 &
  65.92 &
  \multicolumn{1}{c|}{77.17} &
  66.64 &
  58.34 &
  71.82 &
  \multicolumn{1}{c|}{81.67} &
  35.94 &
  28.49 &
  36.84 &
  53.75 \\
\multicolumn{1}{l|}{TeRo (concurrent work)} &
  56.2 &
  46.8 &
  62.1 &
  \multicolumn{1}{c|}{73.2} &
  58.6 &
  46.9 &
  66.8 &
  \multicolumn{1}{c|}{79.5} &
  - &
  - &
  - &
  - \\
\multicolumn{1}{l|}{TeMP-SA (concurrent work)} &
  60.7 &
  48.4 &
  68.4 &
  \multicolumn{1}{c|}{84.0} &
  68.0 &
  55.3&
  76.9 &
  \multicolumn{1}{c|}{91.3} &
  - &
  - &
  - &
  - \\ \hline
\multicolumn{1}{l|}{ChronoR (k=3)} &
  59.39 &
  49.64 &
  65.40 &
  \multicolumn{1}{c|}{77.30} &
  68.41 &
  61.06 &
  73.01 &
  \multicolumn{1}{c|}{82.13} &
  36.50 &
  29.16 &
  37.63 &
  53.53 \\
\multicolumn{1}{l|}{ChronoR (k=2)} &
  62.53 &
  54.67 &
  66.88 &
  \multicolumn{1}{c|}{77.31} &
  67.50 &
  59.63 &
  72.29 &
  \multicolumn{1}{c|}{{82.03}} &
  36.62 &
  29.18 &
  37.92 &
  53.79 \\ \hline
\end{tabular}
}
\caption[Evaluation on the YAGO15k, ICEWS14, and ICEWS05-15 datasets.]{Evaluation on the YAGO15k, ICEWS14, and ICEWS05-15 datasets. Results reported for previous related works are the best numbers reported in their respective paper\footnotemark. }
\label{tab:results}
\end{table*}

Having an appropriate scoring function, one can model the likelihood of any $t_i \in \mathcal{E}$, correctly answering the query $g(h, r, ?, \tau)$ as:

\begin{equation}
\label{eq:likelihood_t_i}
    P(t=t_i\mid h, r, \tau) = \frac{\operatorname{exp}(g(h, r, t_i, \tau))}{\sum_{k=1}^K \operatorname{exp}(g(h, r, t_k, \tau))}
\end{equation}

\noindent and similarly for $P(h=h_i\mid r, t, \tau)$. 

To learn appropriate model parameters, one can minimize, for each quadruple in the training set, the negative log-likelihood of correct prediction:
\begin{align}
\label{eq:loss_base}
    L(\mathcal{K};\bm{\theta}) = \mathlarger{\sum}_{(h, r, t, \tau) \in \operatorname{\mathcal{K}}} & \left( \sum_{t_i, h_i \in \mathcal{E}} -\log(P_{t_{\scriptscriptstyle i}}) -\log(P_{h_{\scriptscriptstyle i}}) \right) 
\end{align}

\noindent where $\bm{\theta}$ represents all the model parameters.

Formulating the loss function following Equation~\ref{eq:loss_base} requires computing the denominator of Equation~\ref{eq:likelihood_t_i} for every fact in the training set of the temporal KG; however it does not require generating negative samples~\cite{armandpour2019robust} and has been shown in practice (if computationally feasible - for our experiments' scale is) to perform better.

\subsection{Regularization}
\label{sec:Regularization}

Various embedding methods use some form of regularization to improve the model's generalizability to unseen facts and prevent from overfitting to the training data. TNTComplEx~\cite{lacroix2020tensor} treats the TKG as an order 3 tensor by \emph{unfolding} the temporal and predicate mode together and adopts the regularization derived for static KGs in ComplEx~\cite{lacroix2018canonical}. Other methods, for example TIMEPLEX~\cite{jain2020temporal}, use a sample-weighted L2 regularization penalty to prevent overfitting.

We also use the tensor nuclear norm due to its connection to tensor nuclear rank~\cite{friedland2018nuclear}. However, we directly consider the TKG as an order 4 tensor with $\bm{E}, \bm{R}, \bm{T}$ containing the rank-one coefficients of its decomposition, and where $\bm{E}, \bm{R}, \bm{T}$ are the tensors containing entity, relation and time embeddings. Based on this connection, we propose the following regularization:

\begin{align}
\label{eq:omega_4}
 \Lambda_4(\bm{\theta}) = \sum_{(h, r, t, \tau) \in \operatorname{\mathcal{K}}} (||\bm{h}||_4^4 + ||\bm{r_2}||_4^4 + ||[\bm{r} | \bm{\tau}]||_4^4 + ||\bm{t}||_4^4)
\end{align}
We empirically compare different regularizations in Section~\ref{sec:Experiments} and show that $\Lambda_4(\bm{\theta})$ outperforms other methods.  We provide the theoretical theorems required to drive Equation~\ref{eq:omega_4} in the Appendix.

\subsection{Temporal Regularization}
In addition, one would like the model to take advantage of the fact that most entities behave smoothly over time. We can capture this smoothness property of real datasets by encouraging the model to learn similar transformations for closer timestamps. Hence, following~\cite{lacroix2020tensor} and~\cite{sarkar2006dynamic}, we add a temporal smoothness objective to our loss function: 

\begin{align}
 \Lambda_\Upgamma = \frac{1}{|\mathcal{T}|-1} \sum_{i=1}^{|\mathcal{T}|-1} || \bm{\tau_{i+1}} - \bm{\tau_{i}}||_4^4
\end{align}
Tuning the hyper-parameter $\lambda_2$ is related to the scale of $\Lambda_\Upgamma$ and other components of the loss function and finding an appropriate $\lambda_2$ can become difficult in practice if each component follows a different scale. Since we are using the 4-norm regularization in $\Lambda_4(\bm{\theta})$, we also use the 4-norm for $\Lambda_\Upgamma$. Similar phenomenal have been previously explored in other domains. For example, in~\cite{belloni2014pivotal} the authors propose $\sqrt{\text{Lasso}}$, where they use the square root of the MSE component to match the scale of the 1-norm used in the sparsity regularization component of Lasso~\cite{tibshirani1996regression} and show improvements in handling the unknown scale in badly behaved systems.

Since we used a 4-norm in $\Lambda_4$, we also use the 4-norm for $\Lambda_\Upgamma$. We saw that in practice using the same order makes it easier to tune the hyperparameters $\lambda_1$ and $\lambda_2$ in $\mathcal{L} (\mathcal{K})$.

\subsection{Loss Function}

To learn the representations for any TKG $\mathcal{K}$, the final training objective is to minimize:

\begin{align}
 \mathcal{L}(\mathcal{K}; \theta) = L(\mathcal{K}; \bm{\theta}) + \lambda_1 \Lambda_4(\bm{\theta}) + \lambda_2 \Lambda_\Upgamma
\end{align}

where the first and the second terms encourage an accurate estimation of the edges in the TKG and the third term incorporates the temporal smoothness behaviour.

\vspace{6pt}

In the next section, we provide empirical experiments and compare ChronoR to various other benchmarks.

\section{Experiments}
\label{sec:Experiments}
\footnotetext{The original results for TNTComplex were reported on the validation set, we use the code and hyper-parameters from the official repository re-run the model and report test set values.}

We evaluate our proposed model for temporal link prediction on temporal knowledge graphs. We tune all the hyper-parameters using a grid search and each dataset's provided validation set. We tune $\lambda_1$ and $\lambda_2$ from $\{10^i| -3 \leq i \leq 1\}$ and the ratio of $\frac{n_r}{n_\tau}$ from $[0.1, 0.9]$ with $0.1$ increments. For a fair comparison, we do not tune the embedding dimension; instead, in each experiment we choose $n$ such that our models have an equal number of parameters to those used in~\cite{lacroix2020tensor}. Table~\ref{tab:chrono_n_size}, in the Appendix, shows the dimensions used by each model for each of the datasets. 

Training was done using mini-batch stochastic gradient descent with AdaGrad and a learning rate of 0.1 with a batch size of 1000 quadruples. We implemented all our models in Pytorch and trained on a single GeForce RTX 2080 GPU. The source code to reproduce the full experimental results will be made public on GitHub.

\subsection{Datasets} \label{sec:Dataset}
We evaluate our model on three popular benchmarks for Temporal Knowledge graph completion, namely ICEWS14, ICEWS05-15, and Yago15K. All datasets contain only positive triples. The first two datasets are subsets of Integrated Crisis Early Warning System (ICEWS), which is a very popular knowledge graph used by the community. ICEWS14 is collected from 01/01/2014 to 12/31/2014, while ICEWS15-05 is the subset occurring between 01/01/2005 and 12/31/2015. Both datasets have timestamps for every fact with a temporal granularity of 24 hours. It is worth mentioning that these datasets are selected such that they only include the most frequently occurring entities (in both head and tail). Below are examples from ICEWS14:

\begin{Verbatim}[fontsize=\small] 

(John Kerry, Praise or endorse, 
        Lawmaker (Iraq), 2014-10-18)
(Iraq, Receive deployment of peacekeepers,
        Iran, 2014-07-05)
(Japan, Engage in negotiation, 
        South Korea, 2014-02-18)
        
\end{Verbatim}

To create YAGO15K, \citet{garcia2018learning}~aligned the entities in FB15K~\cite{bordes2013translating} with those from YAGO, which contains temporal information. The final dataset is the result of all facts with successful alignment. It is worth noting that since YAGO does not have temporal information for all facts, this dataset is also temporally incomplete and more challenging. Below are examples from this dataset\footnote{Some strings shortened due to space.}:

\begin{Verbatim}[fontsize=\small]

(David_Beckham, isAffiliatedTo, Man_U)
(David_B, isAffiliatedTo, Paris_SG)
(David_B, isMarriedTo, Victoria_Beckham, 
                 occursSince, "1999-##-##") 
                 
\end{Verbatim}

\begin{table}[h]
\begin{tabular}{l|ccc}
\hline
          & ICEWS14 & ICEWS05-15  & YAGO15k     \\ \hline
Entities   & 7,128   & 10,488      & 15,403      \\
Relations  & 230     & 251         & 34          \\
Timestamps & 365     & 4,017       & 198         \\
Facts      & 90,730  & 479,329     & 138,056     \\
Time Span  & 2014    & 2005 - 2015 & 1513 - 2017 \\ \hline
\end{tabular}
\caption{Statistics for the various experimental datasets.}
\label{tab:stats}
\end{table}

\begin{table}[h]
\begin{tabular}{l|ccc}
\hline
              & ICEWS14 & ICEWS05-15  & YAGO15k  \\ \hline
ChronoR k=2   & 1600   & 1350        & 1900      \\
ChronoR k=3   & 800    & 700         & 950       \\ \hline
\end{tabular}
\caption{The embedding dimension (n) for each dataset used in our experiments.}
\label{tab:chrono_n_size}
\end{table}

To adapt YAGO15 to our model, following~\cite{lacroix2020tensor}, for each fact we group the relations \textit{occureSince}/\textit{occureUntil} together, in turn doubling our relation size. Note that this does not effect the evaluation protocol. Table~\ref{tab:stats}, summarizes the statistics of used temporal KG benchmarks.

\begin{figure}[h]
    \centering
    \includegraphics[width=0.45\textwidth]{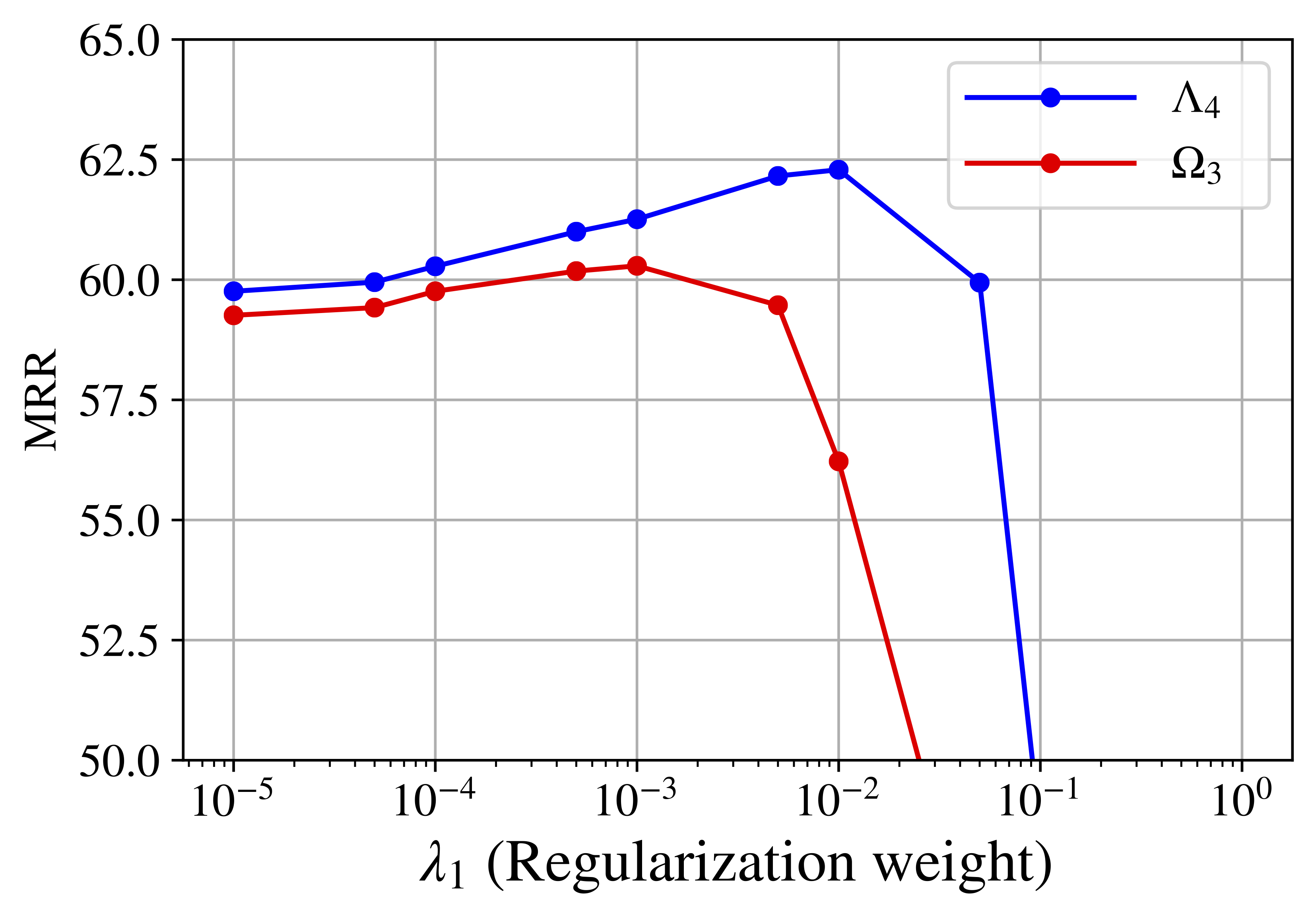}
    \caption{Comparison of various regularizers with different weights on a ChronoR(k=2) trained on ICEWS14.}
    \label{fig:reg_comparison}
\end{figure}

\subsection{Evaluation Metrics and Baselines}

We follow the experimental set-up described in~\cite{garcia2018learning} and~\cite{Goel2020DiachronicEF}. For each quadruple $(h,r,t,\tau)$ in the test set, we fill $(h,r,?,\tau)$ and $(?,r,t,\tau)$ by scoring and sorting all possible entities in $\mathscr{E}$. We report Hits@k for $k=1, 3, 10$ and filtered Mean Reciprocal Rank (MRR) for all datasets. Please see~\cite{nickel2016holographic} for more details about filtered MRR. 

We use baselines from both static and temporal KG embedding models. From the static
KG embedding models, we use TransE, DistMult, SimplE, and ComplEx. These models ignore the timing information. It is worth noting that when evaluating these models on temporal KGs in the filtered setting, for each test quadruple, one must filter previously seen entities according to the fact and its time stamp, for a fair comparison.

To the best of our knowledge, we compare against every previously published temporal KG embedding models that have been evaluated on these datasets, which we discussed the details of in Section~\ref{sec:Related Work}.

\subsection{Results} \label{sec:Results}

In this section we analyze and perform a quantitative comparison of our model and previous state-of-the-art ones. We also experimentally verify the advantage of using Equation~\ref{eq:omega_4} for learning temporal embeddings.

Table~\ref{tab:results} demonstrates link prediction performance comparison on all datasets. ChronoR consistently outperforms all competitors in terms of link prediction MRR and is greater than or equal to the previous work in terms of Hits@10 metric. 

Our experiments with rotations in 3-dimensions show an improvement over ICEWS05-15, but lower performances compared to planar rotations on the other two datasets. We believe this is due to the more complex nature of this dataset (the higher number of relations and timestamps) compared to YAGO15K and ICEWS14. We do not see any significant gain on these three datasets using higher dimensional rotations. Similar to the observations in some static KG benchmarks~\cite{toutanova2015observed}, this might suggest the need for more sophisticated detests. However, we leave further studying of these datasets for future work.

In Figure~\ref{fig:reg_comparison}, we plot a detailed comparison of our proposed regularizer to $\Omega_3$, the regularizer used in TNTComplEx~\cite{lacroix2020tensor}. $\Omega_3$, is a variational form of the nuclear 3-norm and is based on folding the TKG (as a 4-tensor) on its relation and temporal axis to get an order 3 tensor.

We drive $\Lambda_4$ by directly linking the scoring function to the 4-tensors factorization and show that it is the natural regularizer to use when penalizing by tensor nuclear norm. Note that $\Lambda_4$ increases MRR by 2 points and carefully selecting regularization weight can increase MRR up to 7 points.

\section{Conclusion}\label{sec:Conclusion}
We propose a novel k-dimensional rotation based embedding model to learn useful representations from temporal Knowledge graphs. Our method takes into account the change in entities and relations with respect to time. The temporal dynamics of both subject and object entities are captured by transforming them in the embedding space through rotation and scaling operations. Our work generalizes and adopts prior rotation based models in static KGs to the temporal domain. Moreover, we highlight and establish previously unexplored connections between prior scoring and regularization functions. Experimentally, we showed that ChronoR provides state-of-the-art performance on a variety of benchmark temporal KGs and that it can model the dynamics of temporal and relational patterns while being very economical in the number of its parameters. In future work, we will investigate combining other geometrical transformations  and rotations and also explore other regularization techniques as well as closely examine the current temporal datasets as discussed in the experiment section.



\section*{Acknowledgments}
This work is partially supported by NSF under IIS Award \#1526753 and DARPA under Award \#FA8750-18-2-0014 (AIDA/GAIA).

\bibliography{sadeghian_tkge}

\end{document}